\documentclass[10pt,journal]{IEEEtran}
\usepackage{graphicx}      
\usepackage{amsmath}           
\usepackage{amssymb}
\usepackage{algorithm}
\usepackage{algpseudocode}
\usepackage{hyperref}
\usepackage{color}
\usepackage{subfigure}
\usepackage{cite}
\usepackage{enumerate}
\usepackage{epstopdf}
\usepackage{url}

\usepackage[latin1]{inputenc}  

\usepackage{amsthm}

\DeclareMathOperator{\E}{E}
\DeclareMathOperator{\Cov}{Cov}
\DeclareMathOperator{\Var}{Var}
\DeclareMathOperator*{\argmin}{arg\,min}

\newcommand{\mbf}[1]{\mathbf{#1}}
\newcommand{\mbs}[1]{\boldsymbol{#1}}
\newcommand{\what}[1]{\widehat{#1}}
\newcommand{\wtilde}[1]{\widetilde{#1}}

\newcommand{\I}{\mbf{I}}
\newcommand{\0}{\mbf{0}}
\newcommand{\x}{\mbf{x}}

\newcommand{\X}{\mbf{X}}
\newcommand{\wmatrix}{\mbf{W}}
\newcommand{\wvec}{\mbf{w}}
\newcommand{\graph}{\mathcal{V}}

\newcommand{\T}{\top}
\newcommand{\eps}{\varepsilon}
\newcommand{\epsv}{\mbs{\varepsilon}}

\begin{document}

\title{Learning Sparse Graphs for Prediction of
  Multivariate Data Processes}
\author{	\IEEEauthorblockN{Arun~Venkitaraman\IEEEauthorrefmark{1} and Dave Zachariah\IEEEauthorrefmark{2}}\\
\thanks{\IEEEauthorrefmark{1}Department of Information Science and Engineering, School of Electrical Engineering and Computer Science,       
	KTH Royal Institute of Technology,  
	SE-100 44 Stockholm, Sweden. Email: arun.venkitaraman@gmail.com.}
\thanks{{\IEEEauthorrefmark{2}Department of Information Technology, Division of Systems and Control, Uppsala University,
		SE- 752 37, Uppsala, Sweden.
		Email: dave.zachariah@it.uu.se.}}
	\thanks{This work has been partly supported by the Swedish Research Council (VR) under contract 621-2014-5874.}}

\maketitle

\begin{abstract} 
	We address the problem of prediction of multivariate
        data process using an underlying graph model. We develop
        a method that learns a sparse partial correlation graph in a
        tuning-free and computationally efficient manner.  Specifically, the graph structure is learned
        recursively without the need for cross-validation or parameter tuning by
        building upon a hyperparameter-free framework. {\color{black} Our approach does not require the graph to be undirected and also accommodates varying noise levels across different nodes.} Experiments using real-world datasets show that the proposed method offers significant performance gains in prediction, in comparison with the graphs frequently associated with these datasets.
\end{abstract}
\begin{center}
	EDICS -- SAS-STAT, DSP-GRAPH
\end{center}

\section{Introduction}

Complex data-generating processes are often described using graph models \cite{Kolaczyk,Barabasi}. In such models, each node represents a component with a signal. Directed links between nodes represent their influence on each other. For example, in the case of sensor networks, a distance-based graph is often used to characterize the underlying process \cite{Shuman}. {\color{black} The estimation of undirected networks in the context of graph-Laplacian matrix has been considered extensively in the study of graph signals \cite{Dong_laplacian,SR4,SR5,SR6,SR7,Chepuri_laplacian}. Structural equation model and kernel-based methods have been employed for discovery of directed networks in the context of  characterization and community analysis \cite{SR1,SR2}.}  {\color{black} The estimated graphs in these works are usually not applied explicitly for prediction over the graph but rather for characterization and in the context of recovering a graph that closely approximated an underlying graph.} 

In this paper, we are interested in graph models that are useful for prediction where the goal is to predict the signal values at a subset of nodes using information from the remaining nodes. 
To address this task, we aim to learn partial correlation graph models from a finite set of training data. Such graphs can be viewed as the minimal-assumption counterparts of conditional independence graphs \cite{probgraphmodels,partialcorrgraph1}. In the special case of Gaussian signals, the latter collapses into the former. {\color{black}  Further, unlike many graph learning approaches which deal with undirected graphs using a graph-Laplacian approach, we do not assume our graph to be undirected.}

As the size of a graph grows, the number of possible links between nodes grows quadratically. Thus to learn all possible links in a graph model requires large quantities of data. In many naturally occuring and human-made processes, however, the signal values at one node can be accurately predicted from a small number of other nodes. That is, there exists a corresponding sparse graph such that the links to each node are few, directly connecting only a small subset of the graph\cite{Sparsegraphical}. {\color{black} Sparse partial correlation based graphs have been considered earlier in the context of community identification and graph recovery \cite{Meinhausen,Peng}. By taking sparsity into account it is possible to learn graph models with good predictive properties from far fewer samples. The methods for learning sparse models are usually posed as optimization problems and face two major challenges here. First, they require several hyperparameters to specify the appropriate sparsity-inducing constraints as shown below. Second, both tuning hyperparameters and solving the associated optimization problem is often computationally intractable and must be repeated each time the training dataset is augmented by a new data snapshot. This usually involves the use of some technique such as grid-search or some criterion such as Bayesian information (BIC) which adds to the computational complexity. We also note that these prior approaches also implicitly assume the noise or innovation variance to be equal across the different nodes of the graph.}

Our contribution is a method for learning sparse partial correlation graph for prediction that achieves three goals:
\begin{itemize}
\item obviates the need to specify and tune large number of hyperparameters,
{\color{black} which in the general case considered herein scales linearly with the number of nodes,}
\item computationally efficient with respect to the training data size: {\color{black}by exploiting its parallel structure the runtime scales {\em linearly} with the number of observations and {\em quadratically} with the number of nodes.}
\item accommodates {\color{black} varying noise levels} across nodes.
\end{itemize}
The resulting prediction properties are demonstrated using real and synthetic datasets. {\color{black} Experiments with diverse real-world datasets show that our approach consistently produces graphs which result in superior prediction performance in comparison with some of the graph structures employed regularly in analysis of these datasets, e.g., geodesic graphs.}

\emph{Reproducible research:} Code for the method is available at \url{https://github.com/dzachariah/sparse-pcg} and 
\url{https://www.researchgate.net/profile/Arun_Venkitaraman}

\section{Problem formulation}
\label{sec:prob_stat}

We consider a weighted directed graph with nodes indexed by set
$\graph=\{1,2,\cdots,P\}$. Let $x_i$ denote
a signal at the $i$th node and the link from node $j$ to $i$ has a
weight $w_{ij}$. The signals from all nodes are collected in a $P$-dimensional vector $\x \sim p_0(\x)$, where $p_0(\x)$ is
an unknown data generating process. We assume that its covariance
matrix is full rank and, without loss of
generality, consider the signals to be zero-mean. 
Next, we define the weights and related graph quantities.

\subsection{Partial correlation graph}

A partial correlation graph is a property of $p_0(\x)$ and can be
derived as follows. Let
\begin{equation}
\begin{split}
\wtilde{x}_{i} &= x_i - \Cov\big[x_i ,\x_{-(i,j)} \big] \Cov\big[
\x_{-(i,j)} \big]^{-1} \x_{-(i,j)} \\
\wtilde{x}_{j} &= x_j - \Cov\big[x_j ,\x_{-(i,j)} \big] \Cov\big[
\x_{-(i,j)} \big]^{-1} \x_{-(i,j)} 
\end{split}
\label{eq:partial}
\end{equation}
denote the innovations at node $i$ and $j$ after partialling out the signals
from all other nodes, contained in vector
$\x_{-(i,j)}$. The weight of the link from node $j$ to node $i$ is then
defined as
\begin{equation}
\boxed{w_{ij} \triangleq \frac{\Cov\big[\wtilde{x}_{i}, \wtilde{x}_{j}
    \big]}{\Var\big[\wtilde{x}_{j} \big]}, }
\label{eq:weight}
\end{equation} 
which quantifies the predictive effect of node $j$ on node $i$. The graph structure is thus encoded in a $P \times P$ weighted
adjacency matrix $\wmatrix$, where the $ij$th element is equal to
$w_{ij}$ and the diagonal elements are all zero. In many
applications, we expect only a few links incoming to node
$i$ to have nonzero weights. 

 {\color{black}
We can write a compact signal representation associated with the graph
by defining a latent variable $\eps_i = x_i -\sum_{j\neq i} w_{ij}x_j$
at node $i$, with variance $\sigma^2_i$. By re-arranging, we can simply write
\begin{equation} 
\x=\wmatrix \x + \epsv
\label{eq:signalmodelmatrix}
\end{equation}
The variable $\eps_i$ is zero mean and uncorrelated with the signal values
on the right-hand side of row $i$ of \eqref{eq:signalmodelmatrix}, i.e., $\E[ \x_{-i} \eps_i] = \0$.
 This is shown by first using the fact that $\wtilde{x}_{j}$ is uncorrelated with
all elements of $\x_{-(i,j)}$ \cite{KailathEtAl2000_linear} so that $\E[
(x_i - \wtilde{x}_i) \wtilde{x}_j] = 0$. Therefore $\E[
x_j \wtilde{x}_j] = \Var[\wtilde{x}_j]$ and
$\E[\eps_i \wtilde{x}_j] = \E[x_i \wtilde{x}_j ] - w_{ij}\Var[ 
\wtilde{x}_j]  =\E[(x_i -\wtilde{x}_i)
\wtilde{x}_j] = 0$ for all $j\neq i$. Consequently,
$\E[\wtilde{\x}_{-i} \eps_i] = \0$. Using the fact that
$\x_{-i}$ is linearly related to its corresponding innovations
$\wtilde{\x}_{-i}$, it follows that $\E[\x_{-i} \eps_i] =
 \0$.}

\subsection{Prediction}

Having defined the weighted graph above, we now turn to the prediction problem.
		Given a signal $\x_0$ from a
                  subset of nodes $\graph_0 \subset \graph$, the
                  problem is to predict unknown signal values at the
                  remaining nodes $\graph_\star$. An natural predictor
                  of $\x_\star$ is:
		\begin{equation}
		\what{\x}_\star(\wmatrix)=\wmatrix_{\star,0}\x_0,
		\label{eq:predictor}
		\end{equation}
		where $\wmatrix_{\star,0}$ denotes the corresponding submatrix of $\wmatrix$ of size $|\graph_\star|\times |\graph_0|$.
We observe that \eqref{eq:predictor} is a function of $\wmatrix$.
Next, we develop a method for learning $\wmatrix$ from a
dataset
$\mathcal{D} = \big\{ \x(1), \x(2), \dots, \x(N) \big\}$,
where $\x(n)$ denotes the $n$th realization from $p_0(\x)$. The
learned graph $\what{\wmatrix}$ is then used to evaluate a predictor \eqref{eq:predictor}.

\section{Learning a sparse graph}

Let $\wvec^\T_i$ be the $i$th row of $\wmatrix$ after removing the
corresponding diagonal element. Then, for snapshot
$n$, the $i$th row of \eqref{eq:signalmodelmatrix} is given by
\begin{equation}
\begin{split}
x_i(n) \: = \: \underbrace{\wvec^\T_i \x_{-i}(n) }_{\triangleq
  \what{x}_i(n; \wvec_i ) } \: + \: \eps_i(n).
\end{split}
\label{eq:rowmodel}
\end{equation}

A natural approach
to learn a sparse graph $\wmatrix$ from $N$ training samples is:
\begin{equation}
\min_{\wmatrix: \;  \|\wvec_i \|_0\leq K_i, \; \forall i}  \quad
\sum_{i=1}^P \sum_{n=1}^N \left|x_i(n)- \what{x}_i(n; \wvec_i ) \right|^2
\label{eq:learning_l0}
\end{equation}
where the constraint  $\|\wvec_i \|_0\leq K_i \ll P$ restricts the
maximum number of directed links to each node. Let learned
weights be denoted as $\wvec^0_i$, then $\what{x}_i(n; \wvec^0_i )$ is
a sparse predictor of $x_i(n)$ which we take as a reference.  While
this learning approach leads to strictly sparse weights, it has has two drawbacks. First, \eqref{eq:learning_l0} is known
to be NP-hard, and hence convex relaxations must be used practice
\cite{Donoho2006}. Second, a user must specify suitable bounds
$\{ K_i \}$. Tractable convex relaxations of \eqref{eq:learning_l0}, such as the
$\ell_1$-penalized \textsc{Lasso} approach \cite{Tibshirani_LASSO,network_filtering} 
\begin{equation}
\min_{\wmatrix}  \quad
\sum_{i=1}^P \sum_{n=1}^N \left|x_i(n)- \what{x}_i(n; \wvec_i ) \right|^2 +
\lambda_i \| \wvec_i \|_1,
\label{eq:learning_l1}
\end{equation}
avoid an explicit choice of $\{ K_i \}$ but must in turn tune a set of
hyperparameters $\{ \lambda_i \}$ since the variances $\sigma^2_i$ are not uniform in general. With the
appropriate choice of $\lambda_i$, the resulting deviation of $\what{x}_i(n; \what{\wvec}_i )$ from $\what{x}_i(n; \wvec^0_i )$ can be bounded
\cite{Buhlmann&VanDeGeer2011_highdim}. Tuning these hyperparameters
with e.g. cross-validation \cite{HastieEtAl2009_elements} is however computationally
intractable, especially when $N$ becomes large. {\color{black} We note
  here that the approach taken by \cite{Meinhausen} and \cite{Peng} in
  the context of graph discovery implicitly assumes that $\sigma^2_i$
  is equal across nodes, which is a more restrictive assumption.}

An alternative approach is to treat  $\wvec_i$ as a
random variable, with an expected value $\0$, prior to observing data
from the $i$th node. Specifically, consider \eqref{eq:rowmodel}
conditioned on data from all other nodes $\mathbf{X}_{-i}=[\x^\T_{-i}(1)\,
\cdots\,\x^\T_{-i}(N)]^\T$ and assume that 
\begin{equation*}
\E[\wvec_i| \X_{-i}] = \0 \quad \text{and} \quad \Cov[\wvec_i| \X_{-i}] = \text{diag}(\mbs{\pi}_i),
\end{equation*}
where $\mbs{\pi}_i$ is a vector of variances. Under this conditional
model, the MSE-optimal estimator of $\wvec_i$ after observing data from $i$th node
$\underline{\x_i} = [x_i(1), \dots, x_i(N)]^\T$ is expressible as \cite{KailathEtAl2000_linear}:
\begin{equation}
\what{\wvec}_i(\mbs{\pi}, \sigma^2) = \left( \X^{\T}_{-i} \X_{-i} + \sigma^2
  \text{diag}(\mbs{\pi})^{-1}  \right)^{-1} \X^{\T}_{-i}\underline{ \x_i}.
\label{eq:mseweights}
\end{equation}
Similar to the Empirical Bayes approach in
\cite{Berger1985_statistical}, the hyperparameters $\mbs{\pi}$ and
$\sigma^2$ for each $i$ can be estimated by fitting the marginalized covariance model $$\Cov[\underline{\x_i} |
\X_{-i}] = \X_{-i}  \text{diag}(\mbs{\pi}) \X^\T_{-i} + \sigma^2 \I_N$$
to the sample covariance $\what{\Cov}[\underline{\x_i} |
\X_{-i}] = \underline{\x_i}\underline{ \x^\T_i}$  \cite{Anderson1989_linear,StoicaEtAl2011_newspectral}. It was shown in \cite{StoicaEtAl2014_weightedspice}, that evaluating \eqref{eq:mseweights} at the estimated hyperparameters is equivalent to solving a convex, weighted square-root \textsc{Lasso} problem:
\begin{equation}
\what{\wvec}_i(\what{\mbs{\pi}},
\what{\sigma}^2) = \argmin_{\wvec_i} \;
\|\underline{\x_i}-\X_{-i}\wvec_i\|_2+ \sum_{j\neq
i} \frac{\| \underline{\x_{j}} \|_2}{\sqrt{N}} |w_{ij}|,
\label{eq:spice}
\end{equation}
This problem (aka. \textsc{Spice}) can be solved recursively with a
runtime that scales as $\mathcal{O}(NP^2)$
\cite{Zachariah&Stoica2015_onlinespice}. Since each $\what{\wvec}_i$ can be computed in parallel, this can be exploited to obtain $\what{\wmatrix}$ in the same runtime order. Moreover, under fairly
general conditions, the deviation of $\what{x}_i(n; \what{\wvec}_i )$
from $\what{x}_i(n; \wvec^0_i )$ is bounded by known constants \cite{ZachariahEtAl2017_online}.

In sum, using the \textsc{Spice} approach \eqref{eq:spice} we learn a
sparse graph $\what{\wmatrix}$ in a statistically motivated and
computationally efficient manner without the need to search for or
tune hyperparameters. Moreover, it accommodates varying noise levels
across nodes.

\section{Experiments}

We apply the learning method to both synthesized and real-world
multivariate data. We use a training set consisting of $N$
samples to learn a sparse graph. Then by evaluating
\eqref{eq:predictor} at
$\what{\wmatrix}$, we perform  prediction on separate
testing sets. 
The performance is quantified using normalized mean-squared
error evaluated over a test set. Specifically, we define the
normalized prediction error as
\begin{equation*}
\mbox{NPE}=\frac{ \E\big[ \| \x_\star - \what{\x}_\star \|^2_2 \big] }{
 \E\big[ \| \x_\star \|^2_2 \big] }.
\end{equation*}
The expectation is calculated by averaging over different data samples. We evaluate the performance as a function of training set size $N$. The data is made zero-mean by subtracting the component-wise mean from the training and testing sets in all the examples considered. 
For the purposes of comparison, we also perform experiments with the least-squares (LS) estimate of $\mathbf{W}$ obtained as follows:
\begin{equation}
\min_{\wmatrix}  \quad
\sum_{i=1}^P \sum_{n=1}^N \left|x_i(n)- \what{x}_i(n; \wvec_i ) \right|^2
\label{eq:learning_l2}
\end{equation}

\subsection{Synthesized graphs}

\begin{figure}[!h]
	\centering
	\includegraphics[width=0.80\columnwidth]{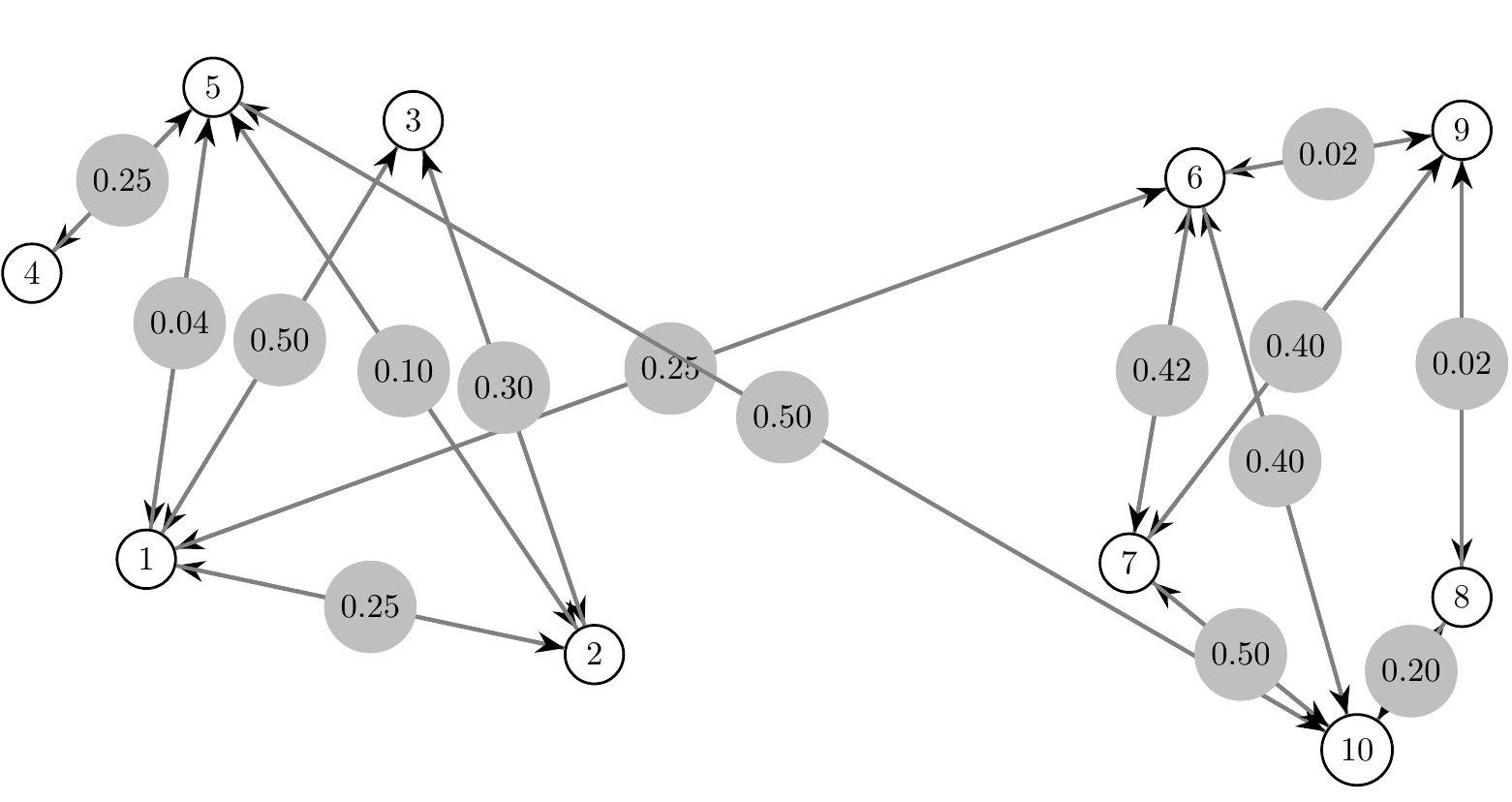}
	\caption{Graph structure with nonzero weights defined by $\wmatrix$.}
	\label{fig:sbm_graph}
\end{figure}
\begin{figure}
	\centering
	\subfigure[]{
		\includegraphics[width=0.4\columnwidth]{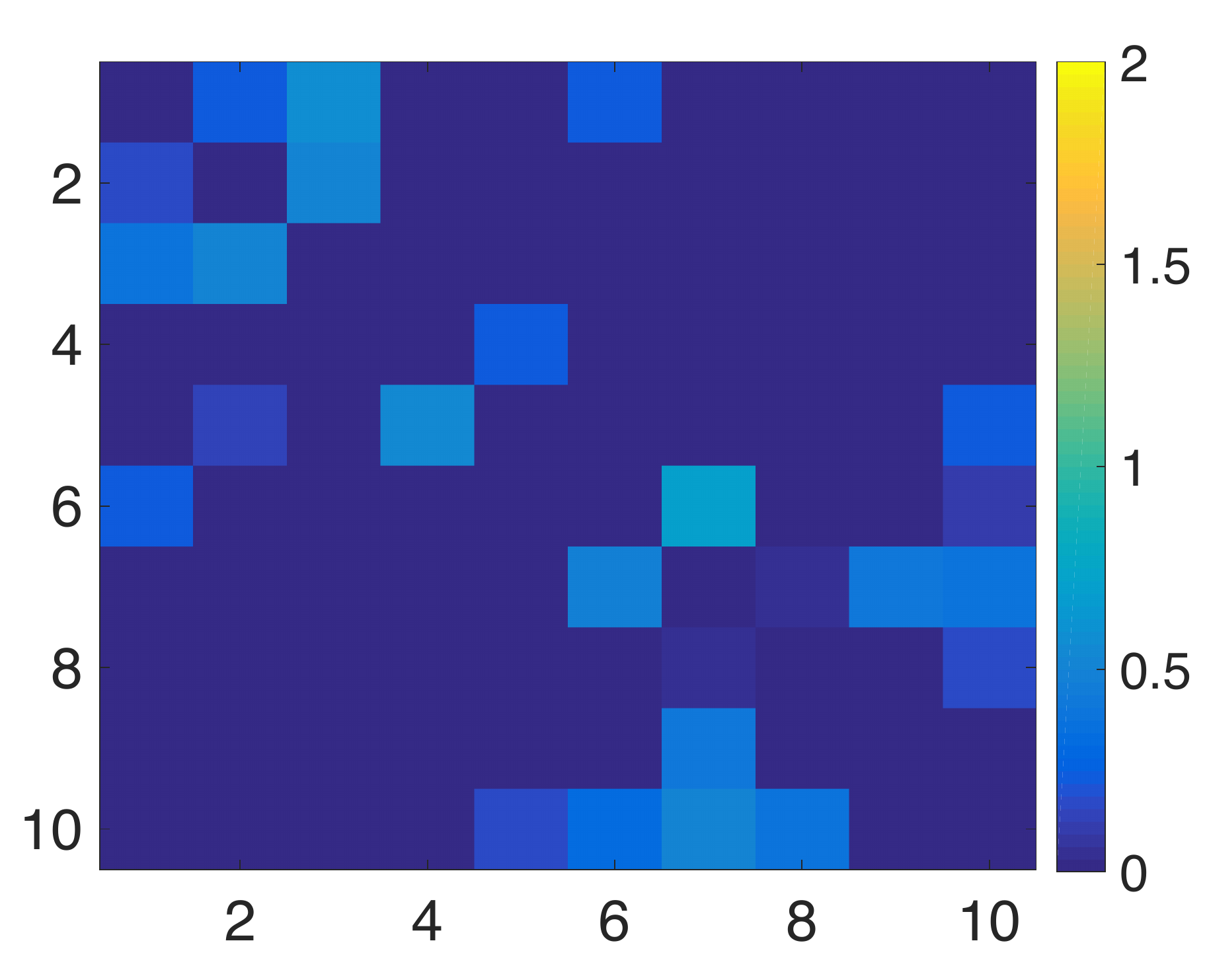}
	}\\
	\subfigure[]{
		\includegraphics[width=0.46\columnwidth]{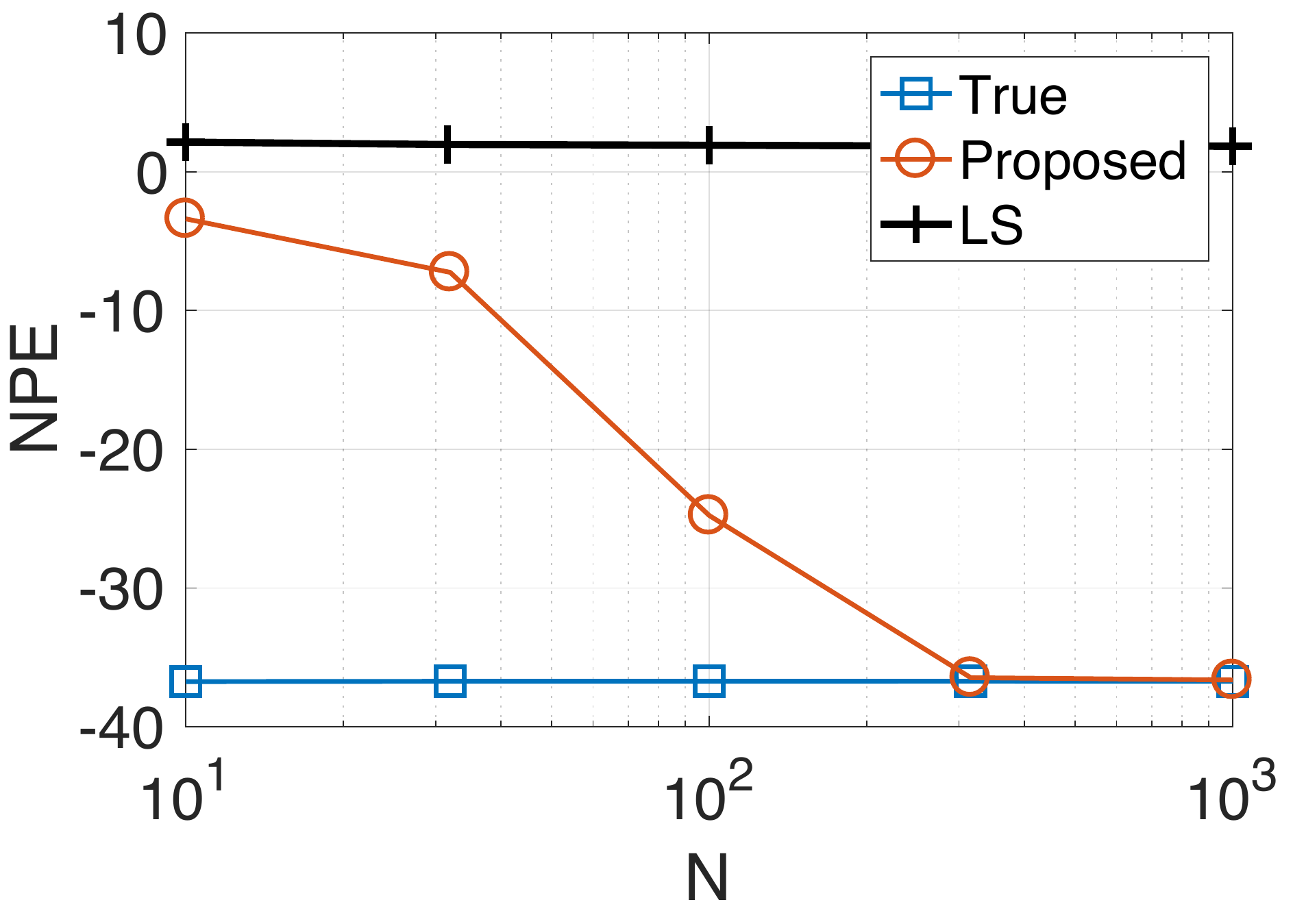}
	}
	\subfigure[]{
		\includegraphics[width=0.46\columnwidth]{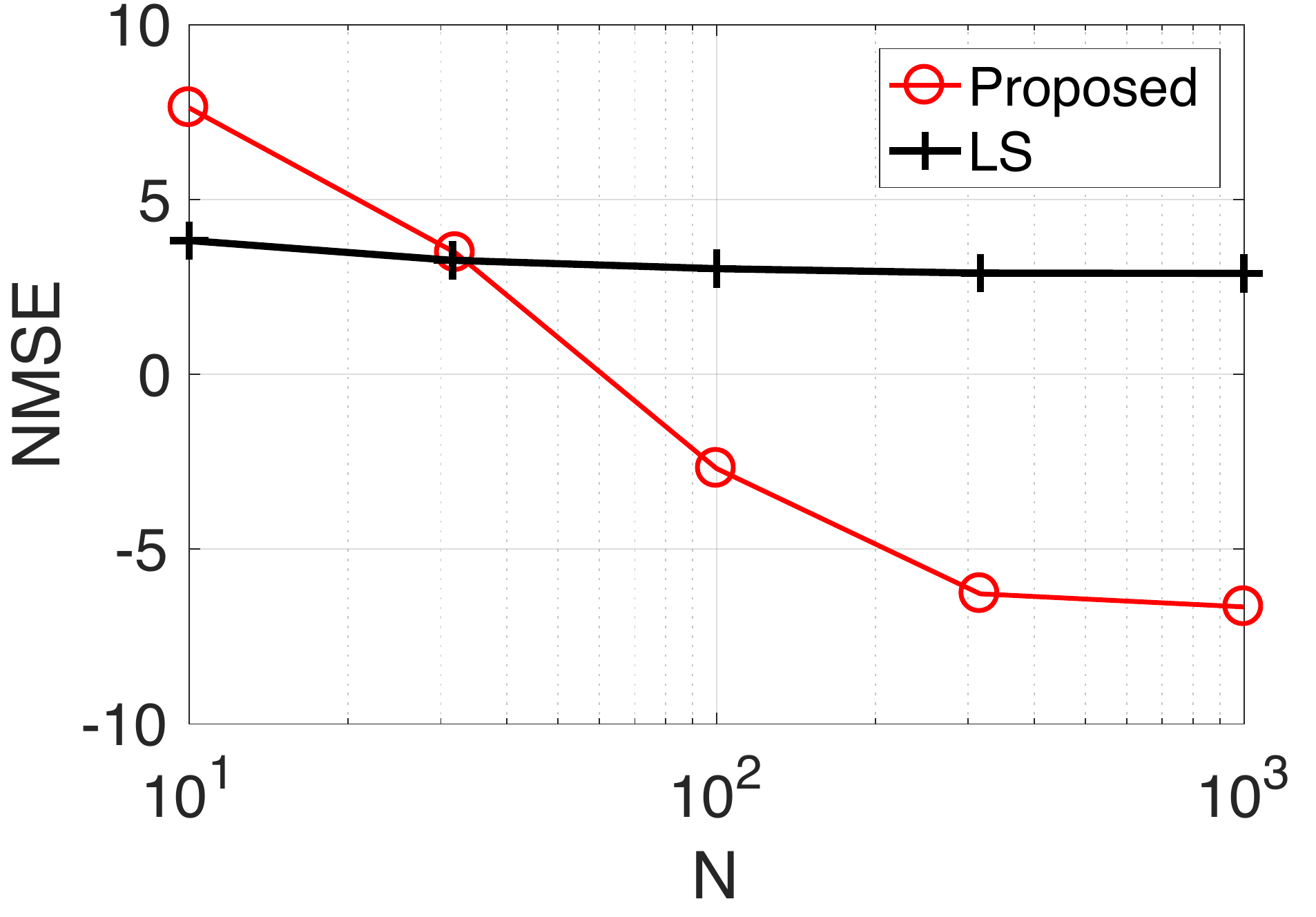}
	}
	\caption{Results for synthesized graph. (a) Example of learned
        graph $\what{\wmatrix}$ using our approach. (b) Prediction errors in [dB], and (c) Normalized error of learned
        graph in [dB].}
	\label{fig:sbm}
\end{figure}

We consider the graph shown in Figure \ref{fig:sbm_graph}
with the indicated edge weights $w_{ij}$ which is akin to a stochastic
block model \cite{Newman}. It consists of two densely connected
communities of 5 nodes each with only two inter-community edges. To
simulate network signals, we generate data as 
\begin{equation}
\x(n) = (\I_P -\wmatrix
)^{-1} \epsv(n),
\end{equation} where the elements of $\epsv(n)$ are mutually
uncorrelated and drawn uniformly from a Gaussian distribution with variances
assigned as $\sigma^2_i \in (0,1]$. We generate a total of
$2 \times 10^4$ samples from which one half is used for training and
remaining for testing by partitioning the total dataset
randomly.  All results are reported by averaging over 500 Monte
Carlo simulations. For sake of illustration, we include an example of $\what{\mathbf{W}}$
from \eqref{eq:spice} when using $N=10^4$ training
samples in Figure~\ref{fig:sbm}(a). 

We perform prediction
experiment using signals at $\graph_0 = \{2,4,6,8,10\}$ to
predict the signals at $\graph_*
 = \{1,3,5,7,9\}$. Figure \ref{fig:sbm}(b) shows that NPE decreases
 with $N$ and ultimately converges to predictions using true $\wmatrix$.
 In Figure~\ref{fig:sbm}(d), we illustrate the rate of overall
 improvement of the learned graph as $N$ increases,
 measured as the normalized MSE $\E[ \|\wmatrix - \what{\wmatrix} \|^2_F]$. We observe that the LS estimator performs poorly in terms of NPE. The NMSE of the LS estimator is also signficantly larger than of our approach. This is expected because the LS estimator is known to exhibit high variance.

\subsection{Flow-cytometry data}
\begin{figure}
	\centering
	\subfigure[]{
		\includegraphics[width=0.38\columnwidth]{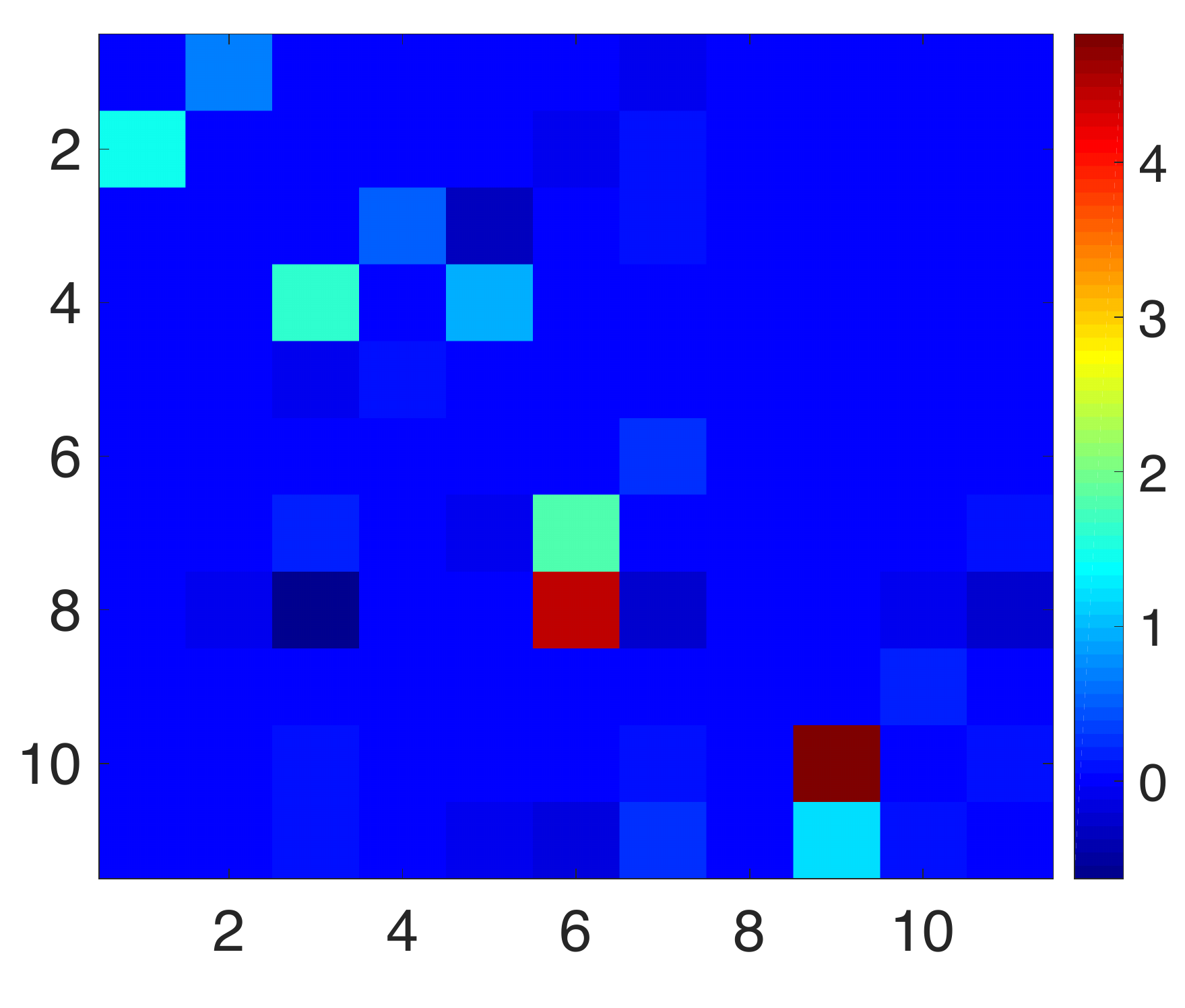}
	}
	\subfigure[]{
		\includegraphics[width=0.46\columnwidth]{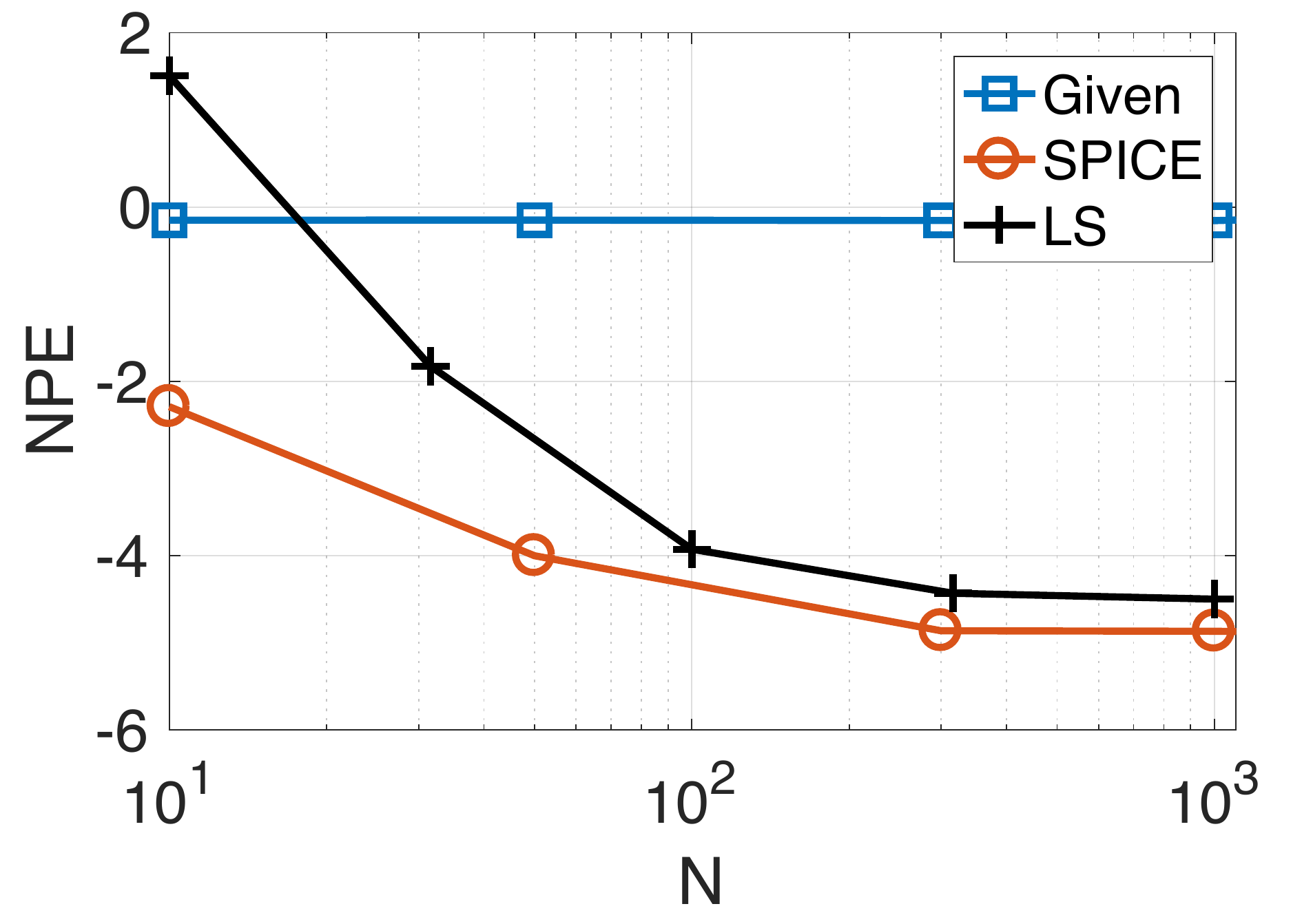}
	}
	\caption{Results for flow-cytometry data: (a) Learned graph
          $\what{\wmatrix}$. (b) (a) Prediction error in [dB].}
	\label{cytometry_fig}
\end{figure}

We next consider flow-cytometry data used in \cite{Sachs}, which
consists of various instances of measurement of  protein and phospholipid components in thousands of
individual primary human immune system cells. The data consists of
total 7200 responses of $P=11$ molecules to different perturbations
which we divide the data into training and test sets. The partition is randomized and for each realization a graph
$\what{\wmatrix}$ is learned. A learned graph is illustrated in
Figure~\ref{cytometry_fig}(a) using $N=3600$ samples. For the prediction
task, we evaluate the performance using 100 Monte Carlo repetitions. For the
sake of comparison, we also evaluate the performance with the sparse binary directed graph
$\wmatrix'$ proposed in \cite{Sachs}. This is because it has been used to encode the directed dependencies between nodes though not specifically designed for
prediction. We
make observations of the signal at nodes $\graph_0=\{3, 8, 9 \}$, noting that these
proteins have the maximum number of connections in
$\wmatrix'$. Prediction is then performed for the signal values at the remaining
nodes $\graph_\star = \{1,2,4 ,5,6,7,10,11\}$. We observe from Figure
\ref{cytometry_fig}(b) that the learned partial correlation graph
$\what{\wmatrix}$ yields superior predictive performance on comparison
with the reference graph $\wmatrix'$. The improvements
saturate beyond $N=10^3$ samples. As expected, the errors of the LS estimator are inflated by its higher variance.

 \subsection{Temperature data for cities in Sweden}
\begin{figure}
	\centering
	\subfigure[]{
		\includegraphics[width=0.46\columnwidth]{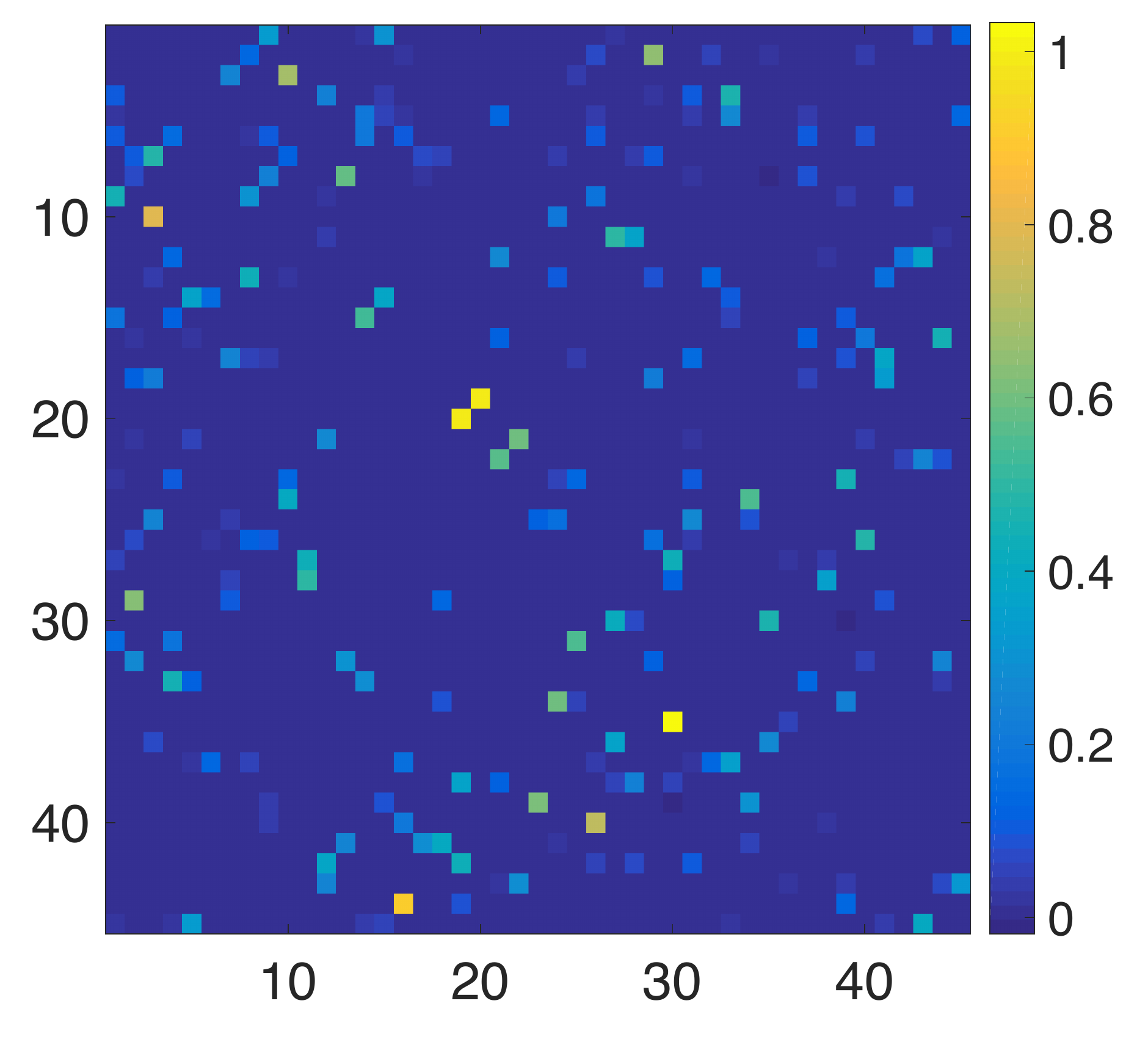}
	}
	\subfigure[]{
		\includegraphics[width=0.46\columnwidth]{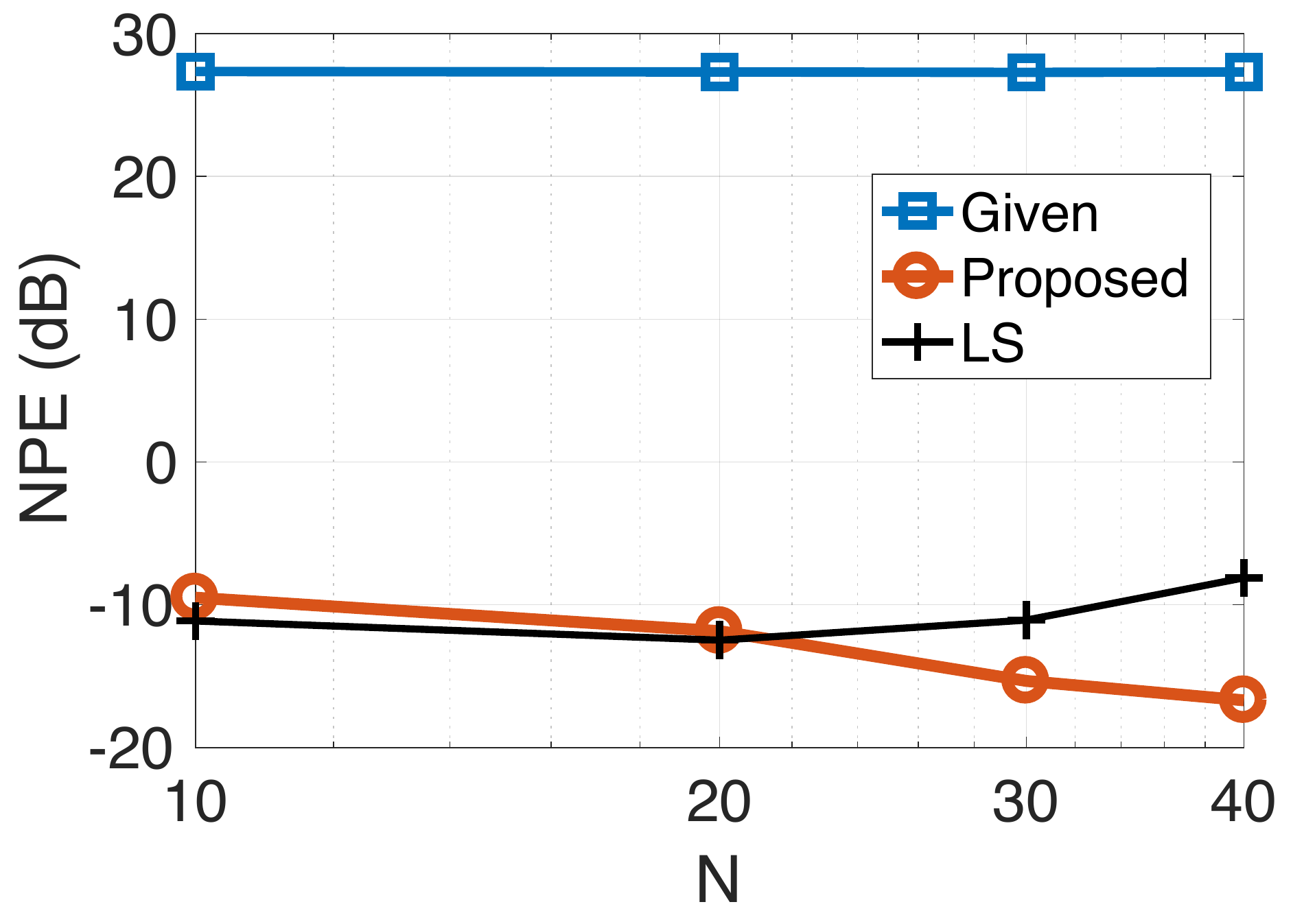}
	}
	\caption{Results for temperature data: (a) Learned graph
          $\what{\wmatrix}$. (b) Prediction error in [dB].}
	\label{temp_fig}
\end{figure}

 We next consider temperature data from the 45 most populated cities
 in Sweden. The data consists of 62 daily temperature readings for the
 period of November to December 2016 obtained from the Swedish
 Meteorological and Hydrological Institute \cite{SMHI}. In
 Figure~\ref{temp_fig}, we show an instance of the learned graph $\what{\mathbf{W}}$ using
 $N=30$ observations. We observe that the graph is sparse as expected.

For the prediction task, we use a subset of $N$ observations for
training and the remaining samples for testing, and perform 100 Monte Carlo repetitions. For reference, we
compare the prediction performance with that of distance-based graph
$\wmatrix'$ with elements $
w'_{ij} = \exp\left(\frac{-d^2_{ij}}{\sum_{i,j}d_{ij}^2}\right)$ for
all  $ i\neq j $ and $0$ for $i=j$. Here $d_{ij}$ denotes the geodesic distance between the $i$th and
 $j$th cities. This graph structure is commonly used in modelling
 relations between data points in spectral graph analysis and in the
 recently popular framework of graph signal processing \cite{Sandry1},
 which makes it a relevant reference. 

The cities are ordered in descending order of their population, and we use
the temperatures of the bottom 40 cities to predict the top 5
cities. That is, $\mathcal{V}_0=\{1,2,3,4,5\}$ and
$\mathcal{V}_\star=\{6,\cdots,45\}$. In Figure~\ref{temp_fig}, we observe
that the prediction performance using the learned graph
$\what{\wmatrix}$ is high already at $N=10$ samples while using
reference graph does not provide meaningful predictions. As with the
earlier examples, we also plot the NPE values obtained for LS. We
observe that the NPE for LS actually increases as the number of
samples is increased. {\color{black}In our learnt graph in
  Figure\ref{temp_fig}(a), the strongest edges are usually across
  cities which are geographically close. Further, a community
  structure is evident between nodes 1 to 15 and between nodes 20 to
  40. This agrees with the observation that the most populated cities
  (nodes 1 to 15) mostly all lie in the south of Sweden and hence, are
  similar in geography and population. Such an observation can also be
  made about the nodes from 20 to 40, since they correspond to the
  relatively colder northern cities.}

 \subsection{64-channel EEG data}
\begin{figure}
	\centering
	\subfigure[]{
		\includegraphics[width=0.46\columnwidth]{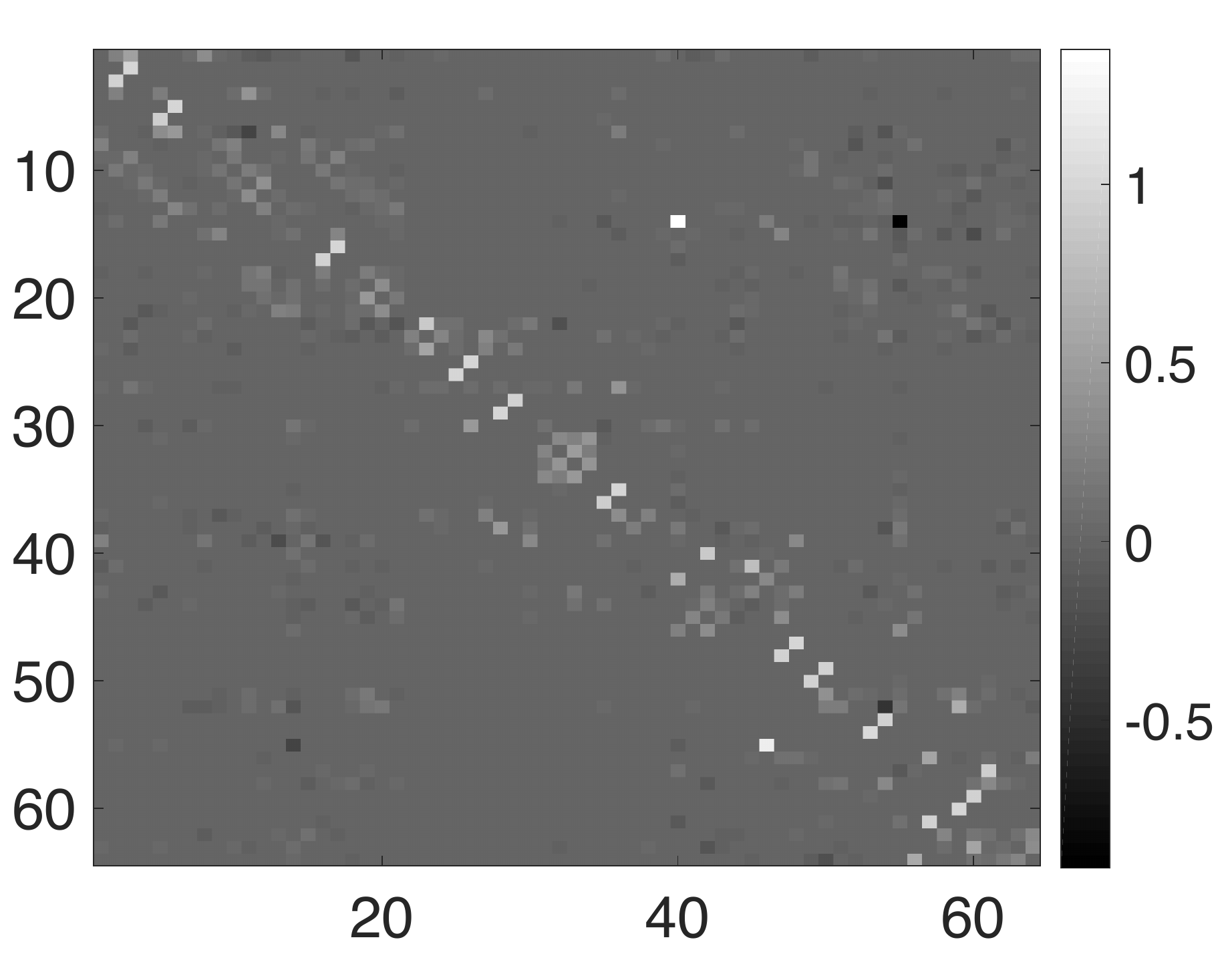}
	}
	\subfigure[]{
		\includegraphics[width=0.46\columnwidth]{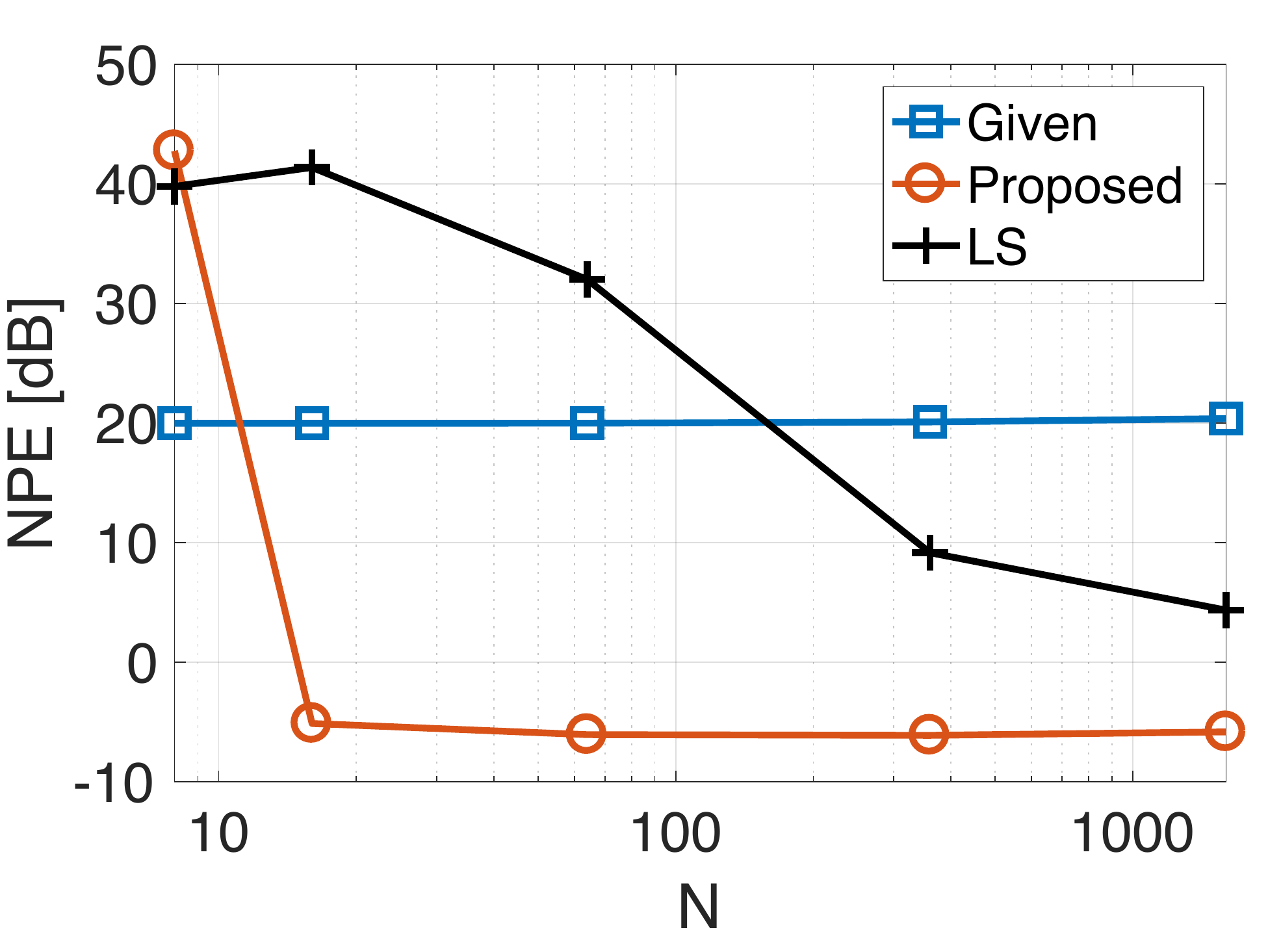}
	}
	\caption{Results for EEG data: (a) Learned graph  $\what{\wmatrix}$. (b) Prediction error in [dB].}
	\label{eeg_fig}
\end{figure}

 Finally, we consider 64-channel electroencephalogram (EEG) signals
 obtained by placing 64 electrodes at various positions on the head of
 a subject and recorded different time-instants \cite{Physionet}. We
 divide the data consisting of 7000 samples into training and test
 sets using 100 Monte Carlo repetitions. An example of a learned graph using our approach is shown in Figure~\ref{eeg_fig}(a). For reference, we compare the prediction performance with that obtained using a
diffusion-based graph $\mathbf{W}'$, where
  $w'_{ij}=\exp\left(-\frac{\|\mathbf{r}_i-\mathbf{r}_j\|_2^2}{\sum_{i,j}\|\mathbf{r}_i-\mathbf{r}_j\|_2^2}\right),\nonumber$
  and $\mathbf{r}_j$ is the vector of 500 successive signal samples
  from a separate set of EEG signals at the $j$th electrode or
  node. In Figure~\ref{eeg_fig}(b) we observe that predictive
  performance using the learned partial
  correlation graph is substantially better than using the
  diffusion-based reference graph and reaches a value close to $-10$dB even at very low training sample sizes. We observe that the NPE with LS estimator remains large even when $N$ is increased.

\section{Conclusion}

We have addressed the problem of prediction of
multivariate data process by defining underlying graph
model. Specifically, we formulated a sparse partial
correlation graph model and associated target quanties for
prediction. The graph structure is learned
recursively without the need for cross-validate or parameter tuning by
building upon a hyperparameter-free framework. Using real-world data
we showed that the learned partial correlation graphs offer superior
prediction performance compared with standard weighted graphs
associated with the datasets. 
\newpage
\bibliographystyle{ieeetr}
\bibliography{refs_graph_learning}

\begin{thebibliography}{10}

\bibitem{Kolaczyk}
E.~D. Kolaczyk, {\em Statistical Analysis of Network Data: Methods and Models}.
\newblock Springer Berlin, 2009.

\bibitem{Barabasi}
A.-L. Barab{\'a}si and M.~P{\'o}sfai, {\em Network Science}.
\newblock Cambridge Univ. Press, 2016.

\bibitem{Shuman}
D.~I. Shuman, S.~Narang, P.~Frossard, A.~Ortega, and P.~Vandergheynst, ``The
  emerging field of signal processing on graphs: Extending high-dimensional
  data analysis to networks and other irregular domains,'' {\em IEEE Signal
  Process. Mag.}, vol.~30, no.~3, pp.~83--98, 2013.

\bibitem{Dong_laplacian}
D.~Xiaowen, T.~Dorina, F.~Pascal, and V.~Pierre, ``Learning laplacian matrix in
  smooth graph signal representations,'' {\em IEEE Trans. Sig. Proc.}, vol.~64,
  pp.~6160--6173, Dec. 2016.

\bibitem{SR4}
D.~Thanou, X.~Dong, D.~Kressner, and P.~Frossard, ``Learning heat diffusion
  graphs,'' {\em IEEE Trans. Signal Info. Process. Networks}, vol.~3,
  pp.~484--499, Sept 2017.

\bibitem{SR5}
S.~Segarra, A.~G. Marques, G.~Mateos, and A.~Ribeiro, ``Network topology
  inference from spectral templates,'' {\em IEEE Trans. Signal Info. Process.
  Networks}, vol.~3, pp.~467--483, Sept 2017.

\bibitem{SR6}
R.~Shafipour, S.~Segarra, A.~G. Marques, and G.~Mateos, ``Network topology
  inference from non-stationary graph signals,'' {\em Proc. IEEE Int. Conf.
  Acoust., Speech Signal Process.}, pp.~5870--5874, March 2017.

\bibitem{SR7}
B.~Pasdeloup, V.~Gripon, G.~Mercier, D.~Pastor, and M.~G. Rabbat,
  ``Characterization and inference of graph diffusion processes from
  observations of stationary signals,'' {\em IEEE Trans. Signal Info. Process.
  Networks}, pp.~1--1, 2017.

\bibitem{Chepuri_laplacian}
S.~P. Chepuri, S.~Liu, G.~Leus, and A.~Hero, ``Learning sparse graphs under
  smoothness prior,'' {\em Proc. IEEE Int. Conf. Acoust., Speech Signal
  Process.}, pp.~6508--6512, 2017.

\bibitem{SR1}
X.~Cai, J.~A. Bazerque, and G.~B. Giannakis, ``Inference of gene regulatory
  networks with sparse structural equation models exploiting genetic
  perturbations,'' {\em PLOS Comput. Biology}, vol.~9, pp.~1--13, 05 2013.

\bibitem{SR2}
Y.~Shen, B.~Baingana, and G.~B. Giannakis, ``Kernel-based structural equation
  models for topology identification of directed networks,'' {\em IEEE Trans.
  Signal Process.}, vol.~65, pp.~2503--2516, May 2017.

\bibitem{probgraphmodels}
D.~Koller and N.~Friedman, {\em Probabilistic Graphical Models}.
\newblock MIT Press, 2009.

\bibitem{partialcorrgraph1}
A.~A. {Amini}, B.~{Aragam}, and Q.~{Zhou}, ``{Partial correlation graphs and
  the neighborhood lattice},'' {\em ArXiv e-prints}, Nov. 2017.

\bibitem{Sparsegraphical}
D.~Angelosante and G.~B. Giannakis, ``Sparse graphical modeling of
  piecewise-stationary time series,'' {\em Proc. IEEE Int. Conf. Acoust.,
  Speech Signal Process.}, pp.~1960--1963., May 2011.

\bibitem{Meinhausen}
N.~Meinshausen and P.~B{\"u}hlmann, ``High-dimensional graphs and variable
  selection with the lasso,'' {\em Annal. Stat.}, vol.~34, no.~3,
  pp.~1436--1462, 2006.

\bibitem{Peng}
J.~Peng, P.~Wang, N.~Zhou, and J.~Zhu, ``Partial correlation estimation by
  joint sparse regression models,'' {\em J. Amer. Stat. Assoc.}, vol.~104,
  no.~486, pp.~735--746, 2009.

\bibitem{KailathEtAl2000_linear}
T.~Kailath, A.~H. Sayed, and B.~Hassibi, {\em Linear estimation}.
\newblock Prentice-Hall, Inc., 2000.

\bibitem{Donoho2006}
D.~L. Donoho, M.~Elad, and V.~N. Temlyakov, ``Stable recovery of sparse
  overcomplete representations in the presence of noise,'' {\em IEEE Trans.
  Inf. Theor.}, vol.~52, pp.~6--18, Jan. 2006.

\bibitem{Tibshirani_LASSO}
R.~Tibshirani, ``Regression shrinkage and selection via the lasso,'' {\em
  Journal of the Royal Statistical Society, Series B}, vol.~58, pp.~267--288,
  1994.

\bibitem{network_filtering}
S.~Yang and E.~D. Kolaczyk, ``Target detection via network filtering,'' {\em
  IEEE Transactions on Information Theory}, vol.~56, pp.~2502--2515, May 2010.

\bibitem{Buhlmann&VanDeGeer2011_highdim}
P.~B{\"u}hlmann and S.~Van De~Geer, {\em Statistics for high-dimensional data:
  methods, theory and applications}.
\newblock Springer Science \& Business Media, 2011.

\bibitem{HastieEtAl2009_elements}
T.~Hastie, R.~Tibshirani, and J.~Friedman, {\em The Elements of Statistical
  Learning: Data Mining, Inference, and Prediction, Second Edition}.
\newblock Springer Series in Statistics, Springer New York, 2009.

\bibitem{Berger1985_statistical}
J.~Berger, {\em Statistical Decision Theory and Bayesian Analysis}.
\newblock Springer Series in Statistics, Springer, 1985.

\bibitem{Anderson1989_linear}
T.~W. Anderson, ``Linear latent variable models and covariance structures,''
  {\em Journal of Econometrics}, vol.~41, no.~1, pp.~91--119, 1989.

\bibitem{StoicaEtAl2011_newspectral}
P.~Stoica, P.~Babu, and J.~Li, ``New method of sparse parameter estimation in
  separable models and its use for spectral analysis of irregularly sampled
  data,'' {\em IEEE Trans. Signal Processing}, vol.~59, no.~1, pp.~35--47,
  2011.

\bibitem{StoicaEtAl2014_weightedspice}
P.~Stoica, D.~Zachariah, and J.~Li, ``Weighted {SPICE}: A unifying approach for
  hyperparameter-free sparse estimation,'' {\em Digital Signal Processing},
  vol.~33, pp.~1--12, 2014.

\bibitem{Zachariah&Stoica2015_onlinespice}
D.~Zachariah and P.~Stoica, ``Online hyperparameter-free sparse estimation
  method,'' {\em IEEE Trans. Signal Processing}, vol.~63, pp.~3348--3359, July
  2015.

\bibitem{ZachariahEtAl2017_online}
D.~Zachariah, P.~Stoica, and T.~B. Sch{\"o}n, ``Online learning for
  distribution-free prediction,'' {\em arXiv preprint arXiv:1703.05060}, 2017.

\bibitem{Newman}
M.~E.~J. Newman, {\em Networks: An Introduction}.
\newblock Oxford University Press, 2010.

\bibitem{Sachs}
K.~Sachs, O.~Perez, D.~Pe{\textquoteright}er, D.~A. Lauffenburger, and G.~P.
  Nolan, ``Causal protein-signaling networks derived from multiparameter
  single-cell data,'' {\em Science}, vol.~308, no.~5721, pp.~523--529, 2005.

\bibitem{SMHI}
``{Swedish Meteorological and Hydrological Institute (SMHI)}.''

\bibitem{Sandry1}
A.~Sandryhaila and J.~M.~F. Moura, ``Discrete signal processing on graphs,''
  {\em IEEE Trans. Signal Process.}, vol.~61, no.~7, pp.~1644--1656, 2013.

\bibitem{Physionet}
A.~L. Goldberger, L.~A.~N. Amaral, L.~Glass, J.~M. Hausdorff, P.~C. Ivanov,
  R.~G. Mark, J.~E. Mietus, G.~B. Moody, C.-K. Peng, and H.~E. Stanley,
  ``Physiobank, physiotoolkit, and physionet,'' vol.~101, no.~23,
  pp.~e215--e220, 2000.

\end{thebibliography}

\end{document}